\documentclass[final,3p,times,twocolumn, nopreprintline]{elsarticle} 

\usepackage{framed,multirow}

\usepackage{amssymb}
\usepackage{latexsym}

\usepackage{url}
\usepackage[dvipsnames]{xcolor}

\usepackage{amsmath,amsfonts,amssymb}
\usepackage{multirow}
\usepackage{tablefootnote}
\usepackage{booktabs}
\usepackage{enumitem}
\usepackage{graphicx}
\usepackage{float}
\usepackage{makecell}
\usepackage{soul}
\usepackage{ulem}
\usepackage{mathtools}
\usepackage{threeparttable}
\usepackage{etoolbox}
\usepackage[numbers]{natbib}

\usepackage[ruled,vlined]{algorithm2e}

\usepackage[bookmarks,bookmarksnumbered]{hyperref}
\hypersetup{pdfauthor=author}

\journal{} 

\begin{document} 

\begin{frontmatter}

\title{TD3Net: A temporal densely connected multi-dilated convolutional network for lipreading}
\author{Byung Hoon Lee \fnref{fn1}}
\author{Wooseok Shin \fnref{fn1}}
\fntext[fn1]{Equal contribution}
\author{Sung Won Han\corref{cor1}} 
\ead{swhan@korea.ac.kr}
\cortext[cor1]{Corresponding author} 
\address{Department of Industrial and Management Engineering, Korea University, Seoul, Republic of Korea}

\begin{abstract}
The word-level lipreading approach typically employs a two-stage framework with separate frontend and backend architectures to model dynamic lip movements. Each component has been extensively studied, and in the backend architecture, temporal convolutional networks (TCNs) have been widely adopted in state-of-the-art methods. Recently, dense skip connections have been introduced in TCNs to mitigate the limited density of the receptive field, thereby improving the modeling of complex temporal representations. However, their performance remains constrained owing to potential information loss regarding the continuous nature of lip movements, caused by blind spots in the receptive field. To address this limitation, we propose TD3Net, a temporal densely connected multi-dilated convolutional network that combines dense skip connections and multi-dilated temporal convolutions as the backend architecture. 
TD3Net covers a wide and dense receptive field without blind spots by applying different dilation factors to skip-connected features.
Experimental results on a word-level lipreading task using two large publicly available datasets, Lip Reading in the Wild (LRW) and LRW-1000, indicate that the proposed method achieves performance comparable to state-of-the-art methods. It achieved higher accuracy with fewer parameters and lower floating-point operations compared to existing TCN-based backend architectures. Moreover, visualization results suggest that our approach effectively utilizes diverse temporal features while preserving temporal continuity, presenting notable advantages in lipreading systems. The code is available at our GitHub repository (https://github.com/Leebh-kor/TD3Net).
\end{abstract}

\begin{keyword} 
Visual speech recognition \sep Lipreading \sep Temporal convolution \sep Dense connectivity \sep Multi-dilation
\end{keyword} 

\end{frontmatter} 

\section{Introduction}
\label{sec:introduction}

Visual speech recognition, also known as lipreading, involves recognizing speech content solely based on lip movements.
Lipreading is beneficial in various scenarios, such as video conferencing in noisy environments, situations requiring silence, and support systems for the hearing impaired, where perceiving utterances from audio is challenging.
Additionally, it can enhance performance in audio-visual speech recognition tasks by providing information that complements the audio signal.
The primary focus of this study is word-level lipreading, which is crucial for recognizing individual words and serves as the foundation for recognizing longer sequences, such as phrases and sentences.

Recently, deep learning-based models have led to significant advancements in lipreading systems, supported by several large-scale datasets \cite{chung2017lip, son2017lip, yang2019lrw}. Furthermore, the performance of lipreading methods has improved with advanced training strategies, such as knowledge distillation (KD) \cite{ma2021towards,ma2022training}, cross-modal memory learning \cite{kim2021multi, kim2022distinguishing}, pretraining on large-scale datasets \cite{ma2022training}, iterative refinement \cite{ryumin2024audio}, and data augmentation  \cite{ma2022training,feng2020learn,zhang2020can}. From the perspective of neural network architectures, a framework comprising frontend and backend architectures has proven to be highly effective for conducting lipreading tasks \cite{assael2016lipnet, son2017lip}. 

The frontend architecture, also known as a visual encoder, processes mouth image frames to extract spatial or spatiotemporal information.
Previous studies \cite{stafylakis2017combining,petridis2018end,martinez2020lipreading,ma2021lip} have used a combination of shallow three-dimensional (3D) convolutional layers and two-dimensional (2D) convolutional networks (e.g., ResNet [15]) as the frontend architecture because they effectively extract spatiotemporal information.
Building on the success of this structure, some researchers have explored a dual-branch structure \cite{wang2019multi, li2022learning} or a hybrid structure that combines 3D convolution and transformers \cite{wang2022lip}.

The backend architecture, which is the focus of our research and is also referred to as a temporal model, learns the temporal dynamics between the features of the sequence received from the frontend phase. Bidirectional recurrent neural networks (BiRNN) have been widely applied to model temporal information among features, focusing on visual representations \cite{petridis2017end, stafylakis2017combining}. Moreover, following demonstrations that temporal convolutional networks (TCNs) \cite{bai2018empirical} outperform recurrent neural network (RNN)-based models in various sequence modeling tasks, some studies \cite{martinez2020lipreading, ma2021lip, kim2022distinguishing} have employed TCNs for lipreading tasks.
A TCN, which is a residual connection-based approach such as ResNet \cite{he2016deep}, uses a one-dimensional (1D) fully dilated convolutional neural network (CNN) to efficiently enlarge the receptive field.
Motivated by the above architecture, \citet{martinez2020lipreading} introduced a multi-scale TCN (MS-TCN) for word-level lipreading, resulting in significant improvements with efficient training costs compared to existing BiRNN-based approaches.
However, an MS-TCN, which is based on sparse connections, cannot compactly reflect temporal features, restricting performance in lipreading tasks that require subtle syllable information to recognize words from frame sequences.
To mitigate these problems, \citet{ma2021lip} proposed a densely connected TCN (DC-TCN) that incorporates dense skip connections \cite{huang2017densely} into a TCN, enabling wide and dense observation of temporal features.

Lipreading involves recognizing utterances by capturing the temporal dynamics of lip movements across consecutive images, such as motion speed, continuity, and transitions. Accordingly, preserving temporal continuity during feature extraction is crucial for accurately modeling correlations between lip features.
However, naively combining dense skip connections with dilated convolutions forces all skip connected features to be processed with the same dilation factor, regardless of their layer depth. This results in blind spots in the receptive field \cite{takahashi2021densely}.
Blind spots refer to discontinuous regions within the receptive field and occur when the receptive field of a feature map---particularly from early layers---is smaller than the applied dilation factor. 
These gaps disrupt the modeling of temporal dependencies, potentially causing the network to overlook critical information about the continuous nature of lip movements.

To address this issue, we propose a temporal densely connected multi-dilated convolutional network (TD3Net) as a novel model for lipreading isolated words.
Our approach is inspired by the densely connected multi-dilated convolutional network (D3Net) \cite{takahashi2021densely}, which was developed for 2D dense prediction tasks, such as image segmentation and audio source separation in the short-time Fourier transform (STFT) domain, with the goal of extracting multi-resolution information and preventing aliasing. TD3Net, which combines dense skip connections and multi-dilated temporal convolutions, covers a wide and dense receptive field without blind spots by applying different dilation factors to the skip-connected features. 
Additionally, a nested dense skip connection structure is utilized to strengthen the modeling of temporal dependencies. This approach enables the efficient learning of multiple temporal features, thereby enhancing the robustness of temporal representations and improving accuracy in utterance discrimination.
To the best of our knowledge, this is the first study to utilize D3Net in the time dimension.

To evaluate the effectiveness of our proposed approach, we conducted experiments using two word-level lipreading benchmark datasets: Lip Reading in the Wild (LRW) \cite{chung2017lip} and LRW-1000 \cite{yang2019lrw}.
Our method consistently achieves performance comparable to state-of-the-art methods across both datasets. Specifically, the experimental results indicate that our approach outperformed representative backend methods---such as the bidirectional gated recurrent unit (BiGRU) \cite{zhao2020mutual}, MS-TCN \cite{martinez2020lipreading}, and DC-TCN \cite{ma2021lip}---while using fewer parameters and requiring less computation under identical settings, except for the backend architecture. Finally, we explain the effectiveness of TD3Net in lipreading systems by analyzing activation maps that reveal how temporal patterns are modeled.

The remainder of this paper is structured as follows: Section \ref{sec:related work} briefly reviews the relevant literature. Section \ref{sec:method} describes the proposed TD3Net architecture in detail. Section \ref{sec:exp} presents the experimental setup and the results obtained. Section \ref{sec:discussion} discusses the issues addressed in this study. Finally, Section \ref{sec:conclusion} summarizes the main conclusions.

\section{Related Work}
\label{sec:related work}
\subsection{Temporal Convolutional Networks}
The RNN-based approaches are used as a standard to model temporal dependencies and have been used in several tasks including lipreading. Alternatively, TCN-based approaches have been employed in tasks involving long sequences, given that the TCN structure is both parallelizable and computationally efficient \cite{oord2016wavenet,dauphin2017language}.
In particular, \citet{bai2018empirical} proposed a generic TCN structure and demonstrated substantial long-term memory capabilities in various sequence modeling tasks, compared to RNN-based models.

To efficiently extract temporal information for lipreading, several studies have employed a TCN-based backend structure \cite{bai2018empirical} as the backend model, which includes two 1D dilated convolutional layers with residual skip connections \cite{he2016deep}.
Among these, MS-TCN \cite{martinez2020lipreading} extended the TCN architecture by introducing a two-stage multi-branch design, where each stage consists of parallel convolutions with different kernel sizes and the same dilation factor. The outputs from the first stage are concatenated along the channel dimension and fed into the second stage, followed by a residual connection that adds the block input to the final output, enabling the model to effectively capture diverse temporal patterns.
Furthermore, DC-TCN \cite{ma2021lip} incorporated dense connections into TCN to more comprehensively model temporal dependencies. In DC-TCN, each layer follows a two-stage multi-branch structure with residual connections, similar to a single MS-TCN block, allowing the model to benefit from both dense and residual connectivity.

TD3Net, which is based on skip connections and 1D dilated convolutions, can be regarded as a TCN-based architecture and as both a simplified and extended version of MS-TCN and DC-TCN.
In particular, we eliminated the intricate components used in MS-TCN and DC-TCN, such as the multi-branch structure and squeeze-and-excitation (SE \cite{hu2018squeeze}) attention blocks.
TD3Net leverages its core architecture without integrating additional complex components, thus maximizing the effectiveness of the multi-dilated TC layers. This simplicity enables easy integration of TD3Net into any lipreading method.

\subsection{Lipreading (Visual Speech Recognition)}
Recent deep learning-based lipreading methods comprise two structures: a visual encoder that extracts spatial information from lip images and a temporal model that extracts temporal information between visual features.

\subsubsection{Frontend -- Visual Encoder}
CNN-based architectures have been adopted in lipreading tasks owing to their success in various image-related domains.
CNN-based visual encoders can be categorized into 2D, 3D, and hybrid 3D + 2D CNN structures \cite{hao2020survey}.
Both the 2D \cite{noda2014lipreading} and 3D CNN \cite{assael2016lipnet} architectures have significant drawbacks despite their notable advancements. While 3D CNNs are computationally intensive because of their 3D operations, 2D CNNs are incapable of extracting spatiotemporal features.
To address these challenges, the community has increasingly adopted mixed models that combine shallow 3D and 2D CNN layers for lipreading tasks \cite{stafylakis2017combining,petridis2018end,martinez2020lipreading,ma2021lip}. The shallow 3D layers extract spatiotemporal information, while the 2D CNN layers focus on spatial features. This mixed approach effectively balances spatiotemporal extraction with manageable computational costs.
In particular, studies utilizing 2D CNN structures \cite{ma2021towards,kim2021multi,kim2022distinguishing,zhang2020can,martinez2020lipreading,ma2021lip,zhao2020mutual} have employed ResNet, a common architecture that has various applications.

Building on this foundation, \citet{wang2019multi} proposed a dual-branch structure comprising two CNN networks (ResNet and Dense3D) along with a fusion module.
Similarly, \citet{li2022learning} proposed the lip slow-fast (LSF) network, derived from the SlowFast network \cite{feichtenhofer2019slowfast}.
Both models utilize each branch to extract distinct representative features, which are subsequently merged by the fusion module to combine diverse feature information. These approaches demonstrated effectiveness in capturing various representations from image frames.
Additionally, \citet{wang2022lip} proposed 3DCvT, which integrates 3D convolution with a transformer architecture \citet{vaswani2017attention} to effectively capture both local and global information from image frames.
Furthermore, by leveraging the success of the EfficientNet architecture  \cite{tan2019efficientnet,tan2021efficientnetv2}, \citet{koumparoulis2022accurate} replaced ResNet with EfficientNet in the 2D CNN backbone, resulting in significant improvements.
More recently, AV-HuBERT \cite{shi2022learning}, a visual encoder that combines ResNet and transformer architectures, has been adopted in various studies \cite{kim2023lip,wang2024whu} owing to its audio-visual pretraining, which facilitates the extraction of visual features that are well-aligned with speech semantics.

This study focuses on developing a novel backend architecture. For the frontend, we employed a standard hybrid encoder that combines shallow 3D CNNs with a 2D ResNet, following common practice in recent lipreading models. Furthermore, similar to a previous study \cite{koumparoulis2022accurate}, we replaced the 2D ResNet in the hybrid encoder with EfficientNetV2 to explore the scalability of the proposed backend architecture.

\subsubsection{Backend -- Temporal Model}
In the temporal model, BiRNN structures comprising two unidirectional RNNs are used to reflect both forward (past-to-future) and backward (future-to-past) temporal representations \cite{petridis2017end,stafylakis2017combining}. Additionally, inspired by the success of TCNs, TCN-based structures have been widely adopted for temporal modeling in lipreading tasks \cite{martinez2020lipreading, ma2021lip, kim2022distinguishing}.
\citet{martinez2020lipreading} expanded the TC blocks \cite{bai2018empirical} into multi-scale TC blocks to aggregate temporal features at different scales. Subsequently, \citet{ma2021lip} argued that MS-TCN models do not cover a dense range of receptive fields and proposed the DC-TCN structure to mitigate this problem. The DC-TCN structure combines dense skip connections with a TCN to comprehensively cover the temporal range of word-level video content. They also adopted the SE attention block to enhance the robustness of temporal representations. However, this approach may create blind spots because it applies a dilation factor that exceeds the temporal coverage of the feature maps \cite{takahashi2021densely}, resulting in the loss of critical temporal information.

Furthermore, several studies \cite{koumparoulis2022accurate,luo2020synchronous,ma2021lira,ryumin2024audio} have used transformer structures to model the temporal relationships between sequence features. Among these, \citet{koumparoulis2022accurate} proposed a novel backend structure that combines a transformer encoder with TCN to exploit the respective strengths of each method.
Moreover, several studies \cite{wang2024whu,ma2021end,chang2024conformer} have adopted the Conformer \cite {gulati2020conformer} architecture, which integrates convolutional modules with self-attention to better capture both local and global temporal dependencies, as an alternative to the standard transformer structure. Although transformer- and Conformer-based backend models improve lipreading performance, their scalability to longer temporal sequences is often limited by quadratic computational complexity. To address this, we propose a novel TCN-based backend architecture that mitigates the limitations of prior TCN-based methods while preserving their computational efficiency.

\subsubsection{Audio-Visual Speech Recognition}
Audio-visual speech recognition (AVSR) has emerged as a prominent research direction, aiming to leverage both visual and audio modalities for more robust speech understanding. This line of research builds upon the progress made in both automatic speech recognition (ASR), which primarily focuses on audio signals, and visual-only lipreading, which seeks to decode speech from lip movements. Building on this foundation, numerous AVSR approaches \cite{ma2021end,burchi2023audio,haliassos2022jointly,yeo2025zero,ma2023auto,rouditchenko2024whisper} have been proposed in recent years, leveraging advances in visual-auditory representation learning.

From an architectural perspective, most AVSR systems still adopt visual processing pipelines similar to those used in conventional lipreading. Specifically, visual features are typically extracted through a frontend encoder, often based on ResNet variants, which are then passed to a temporal backend model that captures sequential dependencies across video frames.
Among various backend architectures, transformer-based models are frequently employed owing to their ability to model long-range temporal dependencies effectively. Additionally, Conformer \cite{gulati2020conformer} and Branchformer \cite{peng2022branchformer} variants have been introduced to improve the modeling of both local and global temporal features, further enhancing the generalization capabilities of audio-visual systems.

Although our study focuses solely on visual input, the proposed TD3Net, an enhanced architecture based on TCN, has the potential to serve as an efficient backend module in multimodal lipreading systems. While transformer-based temporal encoders are widely used for their scalability and ability to model long-range dependencies, their self-attention mechanism incurs quadratic computational complexity with respect to sequence length. This limitation may hinder their practicality in real-time or low-resource scenarios, particularly when processing long video sequences.
By contrast, the proposed TD3Net is inherently based on the TCN architecture, characterized by low computational complexity and a causal design, enabling both low-latency inference and real-time applicability. These properties make it a strong candidate not only as a lightweight alternative but also as a complementary module to transformer-based backend architectures.
Accordingly, the architectural design of TD3Net indicates its broader applicability to future multimodal lipreading frameworks, where efficient temporal modeling remains a critical component.

\subsection{Densely Connected Network}
\begin{figure}
\centering
\centerline{\includegraphics[scale = 0.6]{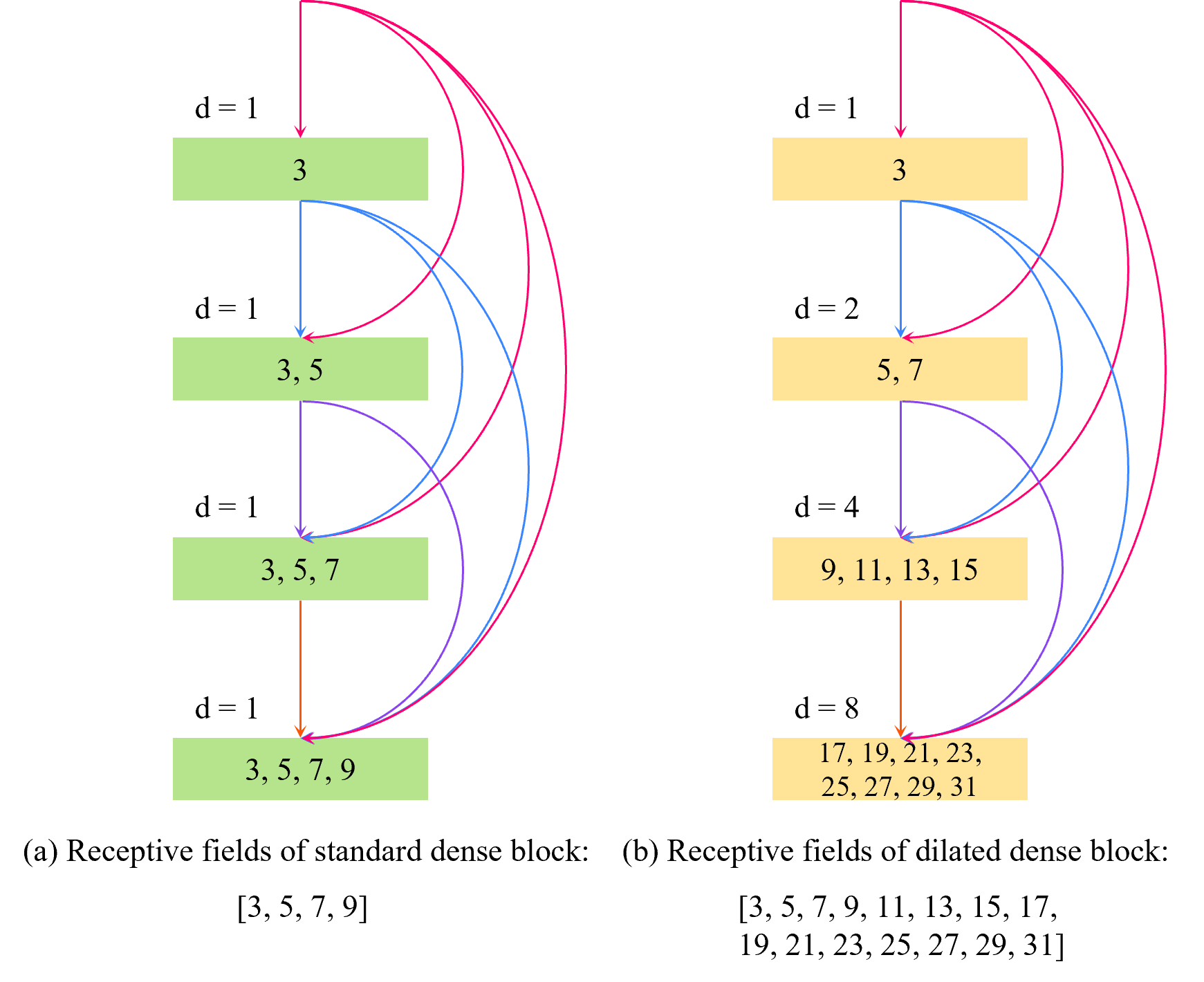}}
\caption{Visualization of the receptive fields for (a) standard dense block and (b) dilated dense block. Each layer is depicted as a square box, with the dilation factor marked as d. The corresponding receptive fields are presented inside the box. For simplicity, the dense blocks comprise a 1D convolutional layer with a filter size of 3 and a stride of 1.}
\label{fig:compare_rf}
\end{figure}

DenseNet \cite{huang2017densely}, in which each layer receives feature maps from all preceding layers, delivers outstanding performance on image classification tasks. This performance is attributed to the dense connectivity pattern, which promotes efficient feature reuse and gradient flow throughout the network, as illustrated in Fig.~\ref{fig:compare_rf}(a).
Moreover, as depicted in Fig.~\ref{fig:compare_rf}(b), using dilated convolution in a dense block enables the network to obtain receptive fields of various scales by applying different dilation factors to the redundant feature maps. 
Several studies have combined dense connectivity and dilated convolution to leverage their advantages. This structure has been widely used in various tasks, such as image super-resolution \cite{shamsolmoali2019single}, speech enhancement \cite{li2019densely}, and semantic segmentation \cite{yang2018denseaspp}.
These structures have also been adopted in sequence modeling tasks such as word-level lipreading \cite{ma2021lip} and sign language translation \cite{guo2019dense} to effectively model subtle temporal variations.

However, as mentioned earlier, naive combinations of dense skip connections and dilated convolutions may create blind spots, leading to an aliasing effect that severely degrades the performance of dense prediction tasks such as image segmentation and audio source separation \cite{takahashi2021densely}. To address this problem, \citet{takahashi2021densely} proposed DenseNet with multi-dilated convolution (D3Net), which performs convolution operations with different dilation factors for each skip connection. 

Although D3Net has demonstrated notable performance in image-based 2D domains, its efficacy in temporal modeling-based 1D domains has not been examined.
To the best of our knowledge, this is the first study to investigate the applicability of D3Net to time-domain tasks.
We modified D3Net to suit temporal modeling while considering both efficiency and performance.
To achieve this, we conducted extensive experiments and used our results to guide network configuration and scalability.
We believe that the proposed TD3Net can benefit various temporal modeling tasks.

\begin{figure}[ht]
\centerline{\includegraphics[scale = 0.68]{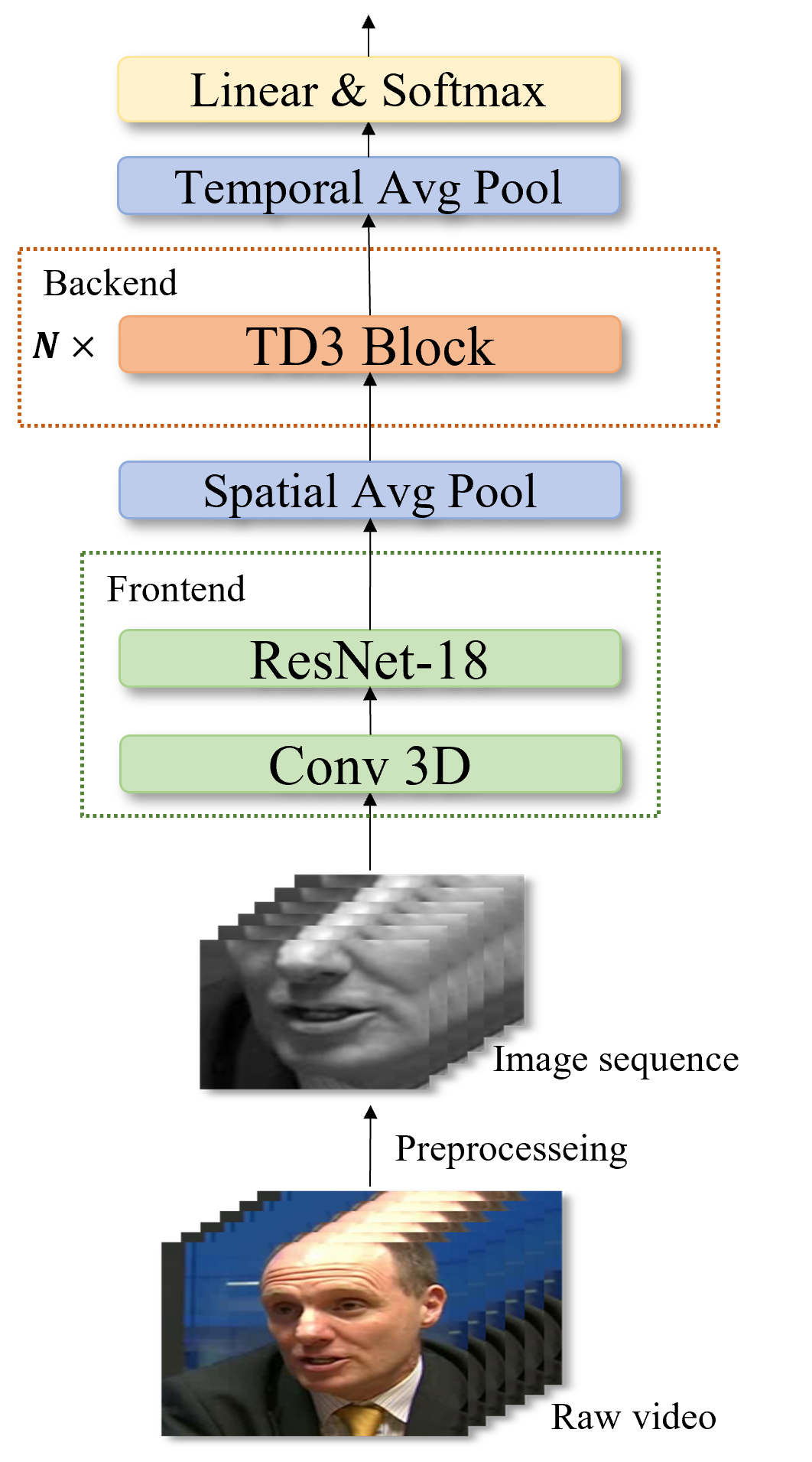}}
\caption{Overall framework of word-level lipreading, including the proposed method.}
\label{TD3Net}
\end{figure}

\section{Method}
\label{sec:method}
As illustrated in Fig.~\ref{TD3Net}, word-level lipreading frameworks primarily comprise two components: frontend and backend. In general, the frontend structure extracts representations that focus on spatial information from sequences of cropped images of the mouth, while the backend structure learns the temporal dynamics between the features of the sequence with fine-grained spatial representations. 
In this section, we first introduce the lipreading framework and then present our proposed model, TD3Net, for the backend phase. Finally, we describe the details pertaining to the proposed method. For a fair comparison, we follow the model architecture adopted in previous studies \cite{ma2021lip}, except for the backend phase.

\begin{figure*}
\centerline{\includegraphics[scale = 0.675]{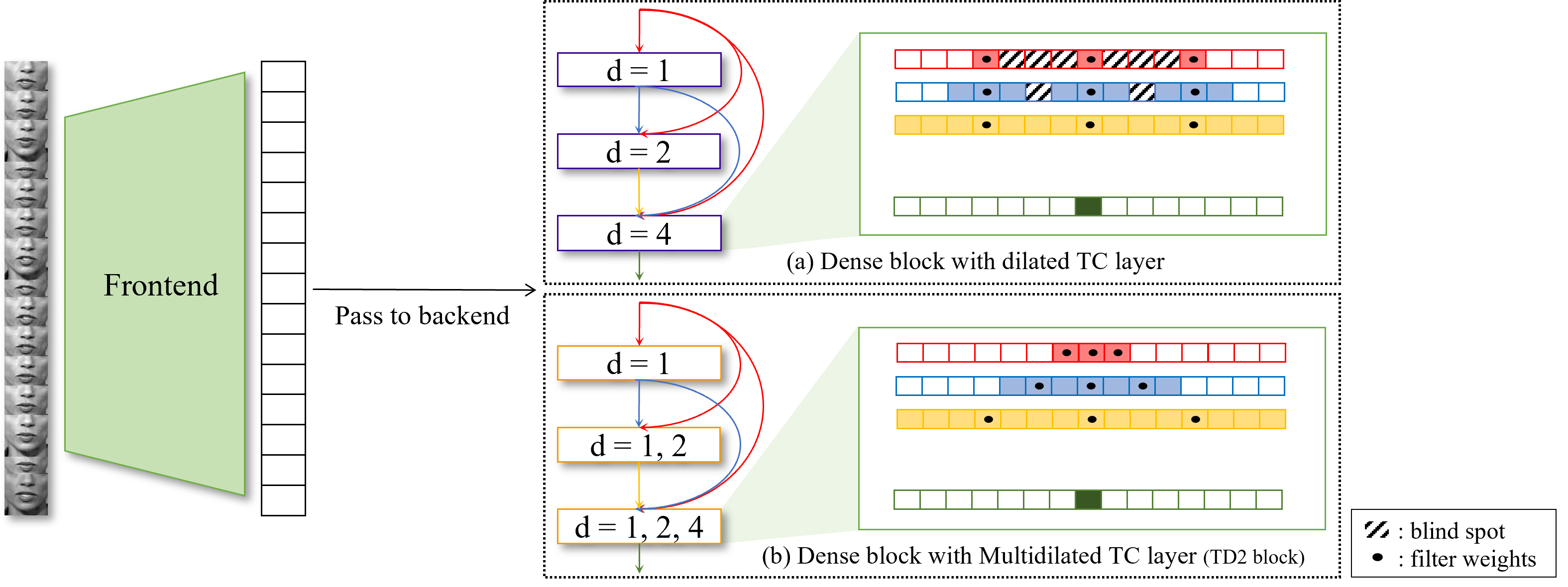}}
\caption{Comparison of blind spots in the receptive field of the green-highlighted output activation in the third layer of the dense block across two TC layer configurations.
In (a), using a fixed dilation factor leads to blind spots when the sampling interval exceeds the receptive fields of activations in the red and blue skip-connected feature maps.
By contrast, (b) illustrates that the multi-dilated TC layer adaptively sets the dilation factor for each skip-connected path based on its receptive field, thereby preventing blind spots and ensuring temporal continuity.}
\label{fig:TMDB}
\end{figure*}

\subsection{Overall Workflow of Word-level Lipreading}
\label{subsec:lip reading framework}
The framework of the proposed method is illustrated in Fig.~\ref{TD3Net}. Given a set of preprocessed (discussed in Section \ref{exp:dataset}) consecutive images of the mouth region $\in \mathbb{R}^{C_{0} \times T \times H_{0} \times W_{0}}$ as the first step of training, shallow 3D convolution extracts the spatiotemporal features $\in \mathbb{R}^{C_{1} \times T \times H_{1} \times W_{1}}$. Here, $C$, $T$, $H$, and $W$ denote the number of channels, sequence length in the temporal dimension, image height, and image width, respectively, and $T$ is maintained during training. Subsequently, by applying 2D ResNet-18 \cite{he2016deep} independently along the temporal dimension, we produce feature maps $\in \mathbb{R}^{C_{2} \times T \times H_{2} \times W_{2}}$, where the spatial representations of sequence features are emphasized. For temporal feature extraction, we aggregate the spatial representations using an average pooling layer to obtain outputs $\in \mathbb{R}^{C_{3} \times T}$. Subsequently, the proposed TD3Net is employed to extract the temporal representations $\in \mathbb{R}^{C_{4} \times T}$ across the temporal sequence. Finally, the class probabilities $\in \mathbb{R}^{C_{5} \times 1}$ are obtained via sequence-wise average pooling, which summarizes the temporal representations, followed by a linear transformation using the weight $\in \mathbb{R}^{C_{4} \times C_{5}}$ and a Softmax layer, where $C_{5}$ denotes the number of target words.

\subsection{Proposed Method}
We propose TD3Net as a variant structure to apply D3Net \cite{takahashi2021densely} as a temporal feature extractor. To directly utilize rich visual information from the frontend phase, we removed the initial two convolutional layers in D3Net, which serve a similar purpose to the stem layer in ResNet.
Before describing the main body of TD3Net, we briefly explain the DenseNet \cite{huang2017densely} architecture as a prerequisite for understanding TD3Net.
DenseNet comprises dense blocks that densely connect convolutional layers, where each layer produces $k$ feature maps as the growth rate. The outputs of the $l$\textsuperscript{th} layer within a block, i.e., $x_{l>0}$, are obtained by using $F_{l}(\cdot)$ on the feature maps in which the outputs of all preceding layers are concatenated as

\begin{equation}\label{eq:dense_block}
  \begin{gathered}
    x_{l} = {F}_{l}([x_{0};x_{1};\cdots;x_{l-1}]),
  \end{gathered} 
\end{equation}
where $F_{l}(\cdot)$ indicates the following $l$\textsuperscript{th} sequential operations: batch normalization (BN), followed by a 2D convolutional layer and nonlinear activation functions. $[x_{0};x_{1};\cdots;x_{l-1}]$ refers to the channel-wise concatenation of the input($x_{0}$) and feature maps received from $1, \cdots , l-1$ layers. Motivated by the aforementioned structures, \citet{ma2021lip} proposed the DC-TCN, which combines dense skip connections with TCN. In the TC layers of a dense block, the dilation factors $d$ are gradually increased as the layers are stacked (e.g., $d$ = 1, 2, 4, 8). 
However, the integration of the TC layers into DenseNet in a naive manner gives rise to blind spots in the receptive field \cite{takahashi2021densely}.

To understand how blind spots arise, it is helpful to revisit the concept of the receptive field.
The receptive field is the input region that influences the computation of an activation at a specific location in the output feature map. Its size and shape are determined by the kernel size, stride, padding, and dilation of convolutional operations, as well as any pooling operations, across all preceding layers in the network.
Since each layer aggregates features from its predecessors, the receptive field progressively expands as the network deepens. With this concept in mind, Fig.~\ref{fig:TMDB}(a) visualizes the receptive field and the blind spots within it for an activation at a specific time step in the output feature map of the third layer of the dense block (highlighted in green). 
This activation is computed via a weighted summation over activations sampled (marked by black dots) from the concatenated skip-connected feature maps using the dilated TC layer.
The colored regions indicate the receptive fields of the sampled activations from each skip-connected feature map, which collectively constitute the receptive field of the green activation; the hatched boxes highlight the blind spots within it.

Although all three feature maps---visualized in red, blue, and yellow---are processed by the same kernel with a dilation factor of 4, the receptive field size of the sampled activations varies with the depth of their originating layers.
In the shallow-path feature maps (indicated by red and blue), each activation has a receptive field narrower than the sampling interval set by the dilation factor. When the kernel samples these activations, the input regions referenced by those activations do not overlap.
Specifically, the sampled activations in shallow-path feature maps reference input regions spanning only 1 and 3 time steps, respectively. When a kernel with a dilation factor of 4 is applied---sampling every 4 time steps---the referenced input regions do not overlap, resulting in disconnected areas, or blind spots, within the receptive field of the output activation. By contrast, activations in the yellow feature map have larger receptive fields, allowing the kernel to sample the temporal axis without information loss. Temporal features extracted from computations involving blind spots within the receptive field could miss critical information about continuous lip movements. Consequently, this may introduce ambiguous representations to the network, potentially causing misinterpretation as different words.

To address this problem, we introduce densely connected multi-dilated TC layers (referred to as the TD2 block in Fig.~\ref{fig:TMDB}(b)). The $l$\textsuperscript{th} multi-dilated TC layer $H_{l}(\cdot)$ of the TD2 block is defined as
\begin{equation}\label{eq:TD2_block}
  \begin{gathered}
    H_{l}(X_{l}) = \sum_{i=0}^{l-1}{d}_{i}\mathrm{TC}^i_{l}(x_{i})\;\in \mathbb{R}^{k \times T},
  \end{gathered}
\end{equation}
where $X_l = [x_0; x_1; \cdots ; x_{l-1}]$ is the input for $l > 0$, and
${d}_{i}\mathrm{TC}^i_{l}(\cdot)$ refers to the subset of the $H_{l}(\cdot)$ applied to $x_{i}$ with a dilation factor $d_{i} = 2^{i}$, producing feature maps with a shape $k \times T$, where $k$ and $T$ denote the growth rate and sequence length, respectively.

As depicted in Fig.~\ref{fig:TMDB}(b), each skip-connected feature $x_i$ is processed individually by a dedicated sub-TC layer of $H_{l}(\cdot)$, with the dilation factor determined by the depth $i$ of the input feature.
The outputs from these sub-TC layers are aggregated through element-wise summation to form the output feature map, where the activation highlighted in green is computed based on the sampled values (marked by black dots).

\begin{figure}
\centering
\centerline{\includegraphics[scale = 0.63]{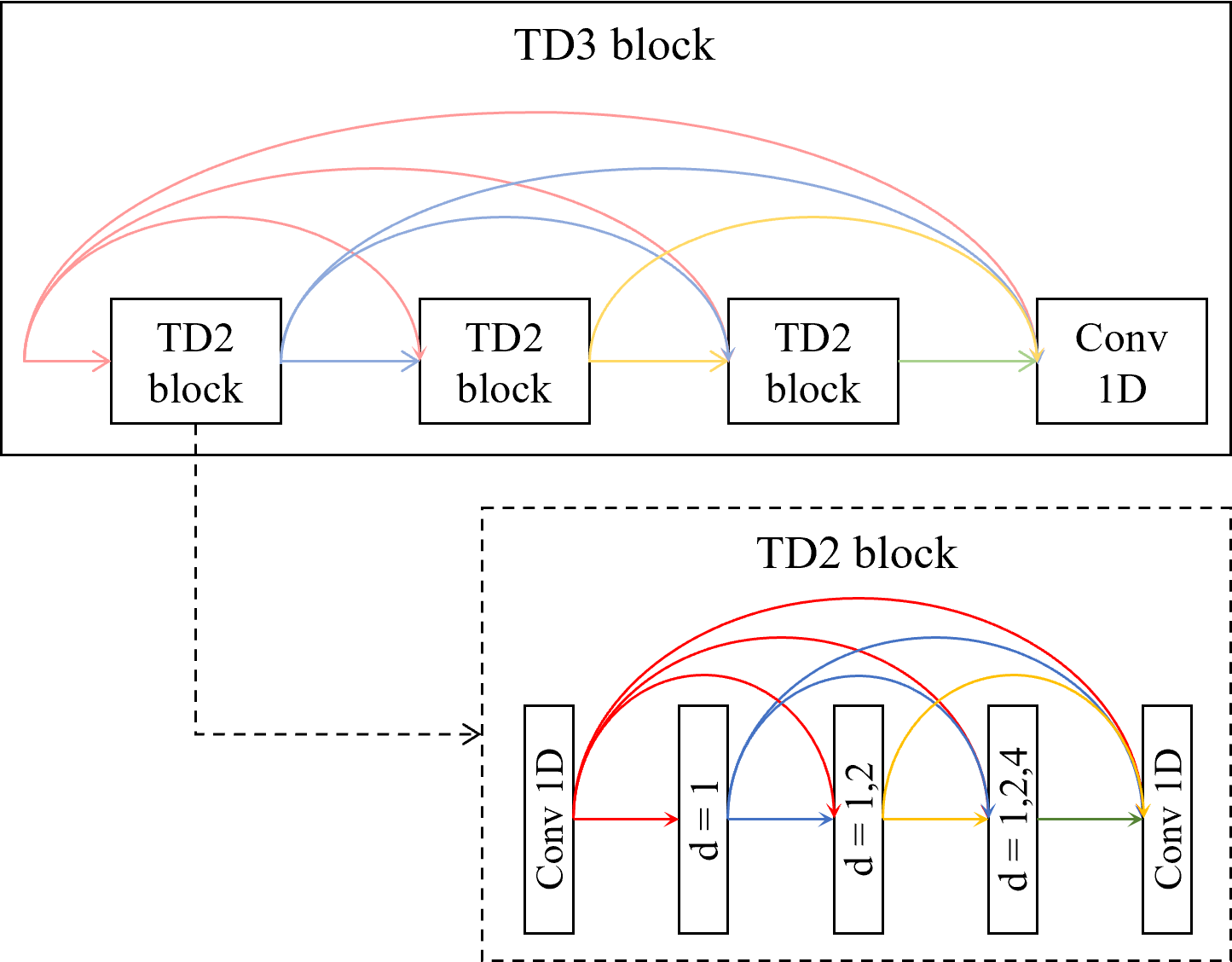}}
\caption{Visualization of the nested structure of the TD3 block that densely connects the TD2 blocks. Each 1D convolution with a kernel size of 1 reduces the number of channels generated by the nested dense skip connections.}
\label{fig:TD3B}
\end{figure}

This multi-dilated design adjusts the dilation factor for each input feature map, ensuring that the sampling interval does not exceed the receptive field of the individual activations within that map. As a result, the receptive fields of the sampled activations are temporally aligned without gaps, allowing all activations in the output feature map to reference continuous input regions without introducing blind spots.
Consequently, the multi-dilated TC layers effectively capture temporal patterns at multiple scales in parallel---smaller dilations capture local dependencies, and larger ones capture long-range dependencies---while maintaining a seamless temporal flow of information. To model temporal dependencies more comprehensively and reuse features across blocks efficiently, multiple TD2 blocks are densely connected, as illustrated in Fig.~\ref{fig:TD3B}. We define this architecture as the TD3 block, and the TD3Net architecture is composed based on these TD3 blocks. 

Furthermore, to mitigate the indiscriminate increase in the number of feature maps resulting from the nested dense skip connections in the TD3 block, we apply a bottleneck layer comprising a 1D convolution with a kernel size of 1. This layer is designed to transform the input and output of each TD2 block and the output from each TD3 block.
The output channels of the TD3 and TD2 blocks are reduced by a compression ratio $c$ and transition ratio $t$, respectively $(0 < c, t < 1)$. In our experiments, the first layer, serving as a bottleneck for each TD2 block, was set to produce $4*k$ feature maps. Notably, TD3Net has several essential parameters, including the number of TD3 blocks $B$, number of TD2 blocks $N$, number of multi-dilated TC layers in a TD2 block $L$, growth rate $k$, and output channel reduction ratios $c$ and $t$. The steps for determining the optimal parameters of the network are described in detail in Sections~\ref{sec:arch_details} and \ref{sec:determining hyperp}. Furthermore, we provide PyTorch-style pseudocode in \ref{sec:appendixa} to clarify the implementation of TD3Net.

\subsection{Architecture Details}
\label{sec:arch_details}
Our network begins with a $3{\times}7{\times}7$ convolutional layer (frame$\times$height$\times$width), followed by batch normalization (BN), ReLU activation, and a $1{\times}3{\times}3$ max pooling layer. This sequence replaces the first convolution and max pooling operations in the standard ResNet-18.
Then, global average pooling is applied at the end of the last convolution, resulting in feature maps with a shape of $512 \times T$ for inputs of shape $1 \times T \times H \times W$, where the leading one indicates a single-channel grayscale image and $T$ (temporal length) is preserved throughout training.
The backend architecture comprises sequential TD3 blocks, each containing an equal number of TD2 blocks. Each TD2 block includes the same number of multi-dilated TC layers with a filter size of 3.
Each multi-dilated TC layer consists of sub-TC layers, where each sub-TC layer comprises a series of TC layers with batch normalization and ReLU activation.
Similar to the transition layer in DenseNet-BC \cite{huang2017densely}, the reduction ratios for output channels $c$ and $t$ are set to 0.5. No channel reduction is applied to the outputs of the last TD3 block.

\section{Experiments}
\label{sec:exp}

\subsection{Dataset}
\label{exp:dataset}
The proposed method was evaluated by conducting experiments on two public benchmark datasets: LRW \cite{chung2017lip} and LRW-1000 \cite{yang2019lrw}. 
These datasets were collected from TV broadcasts, primarily news and talk shows, representing real-world environments rather than controlled laboratory conditions such as the GRID \cite{cooke2006audio} and CUAVE datasets \cite{patterson2002cuave}. 
The LRW and LRW-1000 datasets contain 25 fps video recordings of a single speaker.
Owing to the real-world nature of the data collection, these datasets exhibit significant variations in speaker conditions, including lighting, age, gender, and posture, making them challenging for analysis. 
In the following subsections, we provide a detailed description of the characteristics and preprocessing steps applied to each dataset.

\begin{figure}
\centering
\centerline{\includegraphics[scale = 0.5]{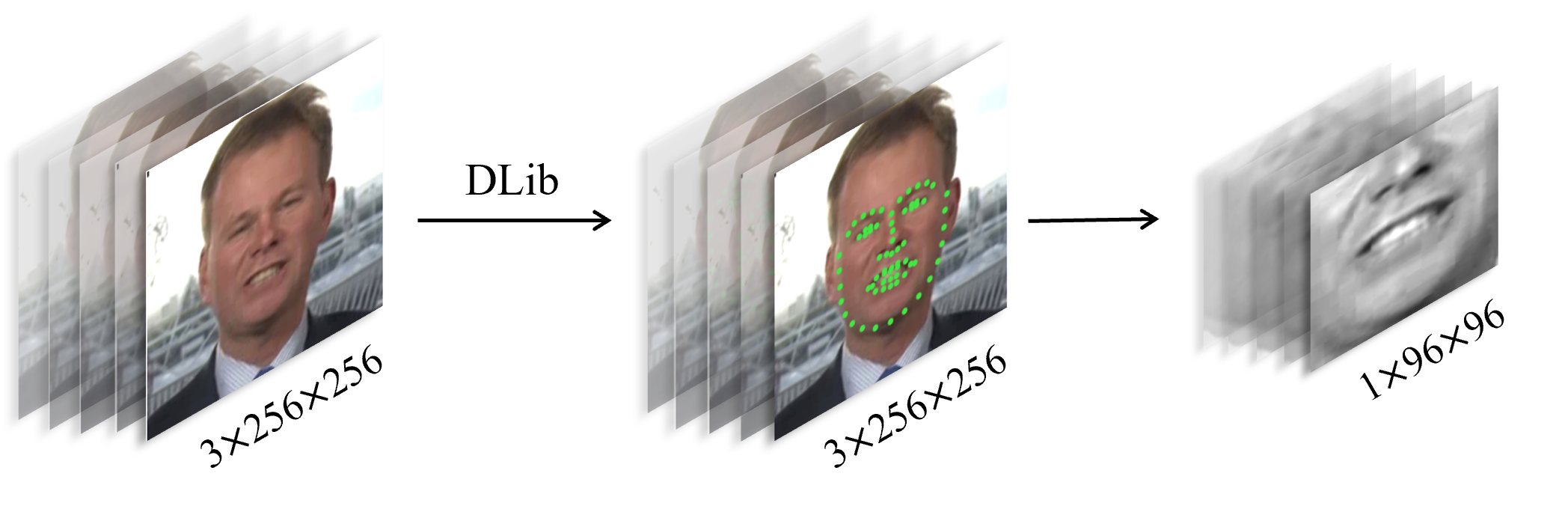}}
\caption{Visualization of the preprocessing steps performed on the LRW dataset, including the image shapes (channels, height, width) at each stage.}
\label{fig:prep}
\end{figure}

\subsubsection{LRW}
LRW is the largest publicly available English word-level dataset and has been widely used in research. The dataset comprises 500 words spoken by more than 1000 speakers, divided into sub-datasets (training, validation, and evaluation sets), comprising 488K, 25K, and 25K videos, respectively. The training set includes 800--1000 utterances for each word, while the validation and evaluation sets contain 50 utterances each. All videos are 1.16 s long (29 frames at a frame rate of 25 fps) and have a fixed resolution of 256$\times$256 pixels. 

In our study, we followed the data preprocessing approach used in previous studies \cite{ma2021lip, martinez2020lipreading}, and Fig.~\ref{fig:prep} illustrates an example of the process. First, we employed the pretrained facial landmark detector from the Dlib toolkit \cite{king2009dlib} to locate 68 key points on the face, including those on the eyes, eyebrows, nose, and mouth. The extracted landmark coordinates enabled us to center the face in each image by aligning it based on these coordinates. Centering the face enhances the performance of lipreading models by emphasizing lip movements, maintaining consistency across frames, and reducing noise by excluding extraneous visual information. Subsequently, we cropped the mouth region from the aligned image to a size of 96$\times$96 pixels and converted it into a grayscale image. This conversion enhanced the robustness of the model to variations in lighting and color while also reducing computational complexity.

\subsubsection{LRW-1000}
LRW-1000 is the largest publicly available Mandarin word-level dataset, comprising 1,000 words spoken by more than 2,000 speakers. The dataset is also pre-partitioned into training, validation, and evaluation sets, containing  603K, 63K, and 52K videos, respectively. Unlike the videos in the LRW dataset, which have fixed resolution and duration, the LRW-1000 dataset comprises multiple videos with varying resolutions and durations. Additionally, it offers a pre-aligned sequence of images in which the lips are centered in each frame of the video, eliminating the need to utilize the facial landmark detector from the Dlib toolkit.

The data were preprocessed according to the pipelines introduced by \citet{feng2020learn}. To ensure consistent video lengths, as in the LRW dataset, all videos were adjusted to 40 frames using zero padding.
Then, each frame, which included an image of the mouth region, was resized to 96$\times$96 pixels and transformed into a grayscale image.

\begin{table*}[t]
\centering
\begin{threeparttable}
\caption{Performance comparison with state-of-the-art word-level lipreading methods on the LRW and LRW-1000 datasets. $WB$ denotes methods using word boundaries as binary indicator vectors. 
Means and standard deviations are reported over three runs.}
\label{tab:sota_comparison}
\small
\setlength{\tabcolsep}{2pt}
\renewcommand{\arraystretch}{0.836}
\begin{tabular}{lllllllll}
\toprule
\multirow{2.5}{*}{Method} && \multicolumn{2}{l}{Backbone} && \multicolumn{2}{l}{Accuracy (\%)} && \multirow{2.5}{*}{$WB$} \\ 
\cmidrule(lr){3-4} \cmidrule(lr){6-7} && Frontend & Backend && LRW & LRW-1000\\
\midrule
\citet{stafylakis2018pushing}  && ResNet 3D & BiLSTM && 88.08 & - && \checkmark \\
\addlinespace
\citet{wang2019multi} / Dual-branch  && ResNet + Dense3D & Bi-ConvLSTM && 83.34 & 36.91 && \\
\addlinespace
\citet{weng2019learning} && Two-Stream ResNet & BiLSTM && 84.07 & - && \\
\addlinespace
\citet{zhao2020mutual} && 3D $+$ ResNet & BiGRU && 84.41 & 38.79 && \\
\addlinespace
\citet{zhang2020can} / Cutout && 3D $+$ ResNet & BiGRU && 85.02 & 45.24 && \\
\addlinespace
\citet{martinez2020lipreading} && 3D $+$ ResNet & MS-TCN && 85.30 & 41.4 && \\
\addlinespace
\citet{feng2020learn} / Training tricks && 3D $+$ SE-ResNet & BiGRU && 88.40 & 55.70 && \checkmark \\
\addlinespace
\citet{ma2021towards} / KD \& Ensemble && 3D $+$ ResNet & MS-TCN && 88.50\tnote{$\dagger$} & 46.60\tnote{$\dagger$} && \\
\addlinespace
\citet{kim2021multi} / Cross-modal memory && 3D $+$ ResNet & BiGRU && 85.40\tnote{*} & 50.82\tnote{*} && \\
\addlinespace
\citet{ma2021lip} && 3D $+$ ResNet & DC-TCN && 88.36 & 43.65 && \\
\addlinespace
\citet{kim2022distinguishing} / Multi-head memory && 3D $+$ ResNet & MS-TCN && 88.50\tnote{*} & 53.82\tnote{*} && \\
\addlinespace
\citet{wang2022lip} && 3DCvT & BiGRU && 88.50 & 57.50 && \checkmark \\
\addlinespace
\citet{li2022learning} / SlowFast && 3D + 2D LSF & BiGRU && 88.52 & 58.17 && \checkmark \\
\addlinespace
\citet{ivanko2022visual} / Preprocessing using audio && 3D + SE-ResNet & BiLSTM && 88.70\tnote{*} & - && \\
\addlinespace
\citet{koumparoulis2022accurate} && 3D + EfficientNetV2-L & Transformer + TCN && 89.52 & - && \\
\addlinespace
\multirow{2}{*}{\citet{ryumin2024audio} / Iterative Refinement} && \multirow{2}{*}{ResNet} & \multirow{2}{*}{BiLSTM + Transformer} && 88.36 & - && \\
& & & & & 89.57\tnote{$\dagger$} & - && \\
\addlinespace
\midrule
\multirow{3}{*}{TD3Net-Best (ours)} && 3D + EfficientNetV2-S & \multirow{3}{*}{TD3Net} && 89.86$\pm0.06$ & 52.01$\pm0.10$ & \\
&& 3D + ResNet &  && 89.54$\pm0.08$ & 51.41$\pm0.07$ && \\
&& 3D + ResNet &  && 91.41$\pm0.10$ & 59.20$\pm0.11$ && \checkmark\\
\bottomrule
\end{tabular}
\begin{tablenotes}
\footnotesize \item[*] Bimodal Training 
\footnotesize \item[$\dagger$] Ensemble
\end{tablenotes}
\end{threeparttable}
\end{table*}

\subsubsection{Word Boundary}
Word boundaries, which indicate the beginning and end of target word utterances, are provided by both datasets. In the early stages, they were used to remove the out-of-boundary frames of the target word \cite{chung2018learning}. As an alternative to removing the frames, \citet{stafylakis2018pushing} added word boundaries to the network as binary indicator vectors that provide contextual information about words. To further enhance the performance of the proposed model, we explored the effect of word boundaries on it.

\begin{table*}[t]
\centering
\setlength{\tabcolsep}{8pt} 
\caption{
Comparison of backend architectures on the LRW test dataset. All models use the same frontend, and only backend parameters and FLOPs are reported to focus the comparison on backend design. Inference time reflects the full model (frontend + backend), measured on the entire test set with a batch size of 32.}
\label{table:main_result}
\begin{tabular}{lllll}
\toprule
Method (Backend) & Backend Params (M) & Backend FLOPs (G) & Inference Time (s) & Accuracy (\%) \\
\midrule
BiGRU$^{\dagger}$ \cite{zhao2020mutual}       & 37.78 & 2.20 & 39 & 85.35 $\pm$ 0.12 \\
MS-TCN$^{\dagger}$ \cite{martinez2020lipreading} & 24.80 & 2.24 & 44 & 86.90 $\pm$ 0.13 \\
DC-TCN$^{\dagger}$ \cite{ma2021lip}           & 40.55 & 2.94 & 54 & 87.92 $\pm$ 0.09 \\
TD3Net-Base                                   & 18.69 & 1.56 & 45 & 89.36 $\pm$ 0.09 \\
TD3Net-Best                                   & 31.89 & 1.92 & 49 & 89.54 $\pm$ 0.08 \\
\bottomrule
\end{tabular}
\end{table*}

\subsection{Training Details}
\label{sec: training details}
We followed a training procedure similar to that described in previous studies \cite{ma2021lip}. The training settings were nearly identical for both the LRW and LRW-1000 datasets, with specific adjustments made to prevent overfitting. A network with randomly initialized weights was trained to recognize words in the video by minimizing the cross-entropy error. For training, we set a batch size of 32 and selected 120 epochs. We used the AdamW optimizer \cite{loshchilov2017decoupled} with a $l_{2}$ weight decay of 1 $\times 10^{-2}$. The learning rate was initialized at 3 $\times 10^{-4}$ and steadily decreased to 5 $\times 10^{-8}$ using a cosine annealing scheduler. The images of the mouth region were randomly cropped to a size of $88{\times}88$, and a horizontal flip was applied with a probability of 0.5 for data augmentation. Referring to the code released by \citet{ma2021lip}, we also included the Mixup \cite{zhang2017mixup} data augmentation technique by setting the value of $\alpha$ to 0.4, where $\alpha$ determined the degree of interpolation between video-word pairs. A dropout \cite{srivastava2014dropout} with a probability of 0.2 for LRW and 0.1 for LRW-1000 was applied to avoid overfitting after each TC layer, which comprised a 1D dilated convolutional layer, BN, and ReLU activation function. For testing, each frame of the video was cropped from the center to $88{\times}88$ pixels.
All experimental results are reported as the mean and standard deviation measured over three runs conducted in the following environments: UBUNTU 20.04, PYTHON 3.8.13, PyTorch 1.8.0, CUDA 11.1, and NVIDIA RTX 3090.

\subsection{Experimental Results}
\label{sec:}
To verify the efficiency of the proposed method, we compared its word-level lipreading performance with that of state-of-the-art models. Table~\ref{tab:sota_comparison} summarizes the word accuracies of TD3Net and other state-of-the-art methods on the LRW and LRW-1000 test datasets.
Under the ResNet frontend settings, the proposed method outperformed existing methods, achieving a word accuracy of $89.54\%$ on the LRW dataset.
Notably, the $89.57\%$ observed in \cite{ryumin2024audio} is comparable to the $89.54\%$ achieved by TD3Net, but it was achieved using a multi-prediction strategy similar to an ensemble, which incurs high inference costs. 
In other words, under single inference conditions, TD3Net outperforms all existing methods.
Furthermore, TD3Net with the EfficientNetV2 frontend significantly outperformed all existing methods, achieving a word accuracy of $89.86\%$ on the LRW dataset. These results indicate that TD3Net functions effectively as a backend model for capturing temporal relationships, regardless of the encoder structure.
On the LRW-1000 dataset, our models achieved competitive results, thereby outperforming other methods without information about word boundaries or modality (audio signal). After incorporating the word boundaries, our method achieved the highest accuracies among all compared methods, delivering outstanding performance on the LRW and LRW-1000 datasets. 

Table~\ref{table:main_result} presents the results of the proposed backend architecture in comparison with three representative alternatives---BiGRU~\cite{zhao2020mutual}, MS-TCN~\cite{martinez2020lipreading}, and DC-TCN~\cite{ma2021lip}---on the LRW test dataset. The evaluation focused on the impact of backend design in terms of recognition accuracy, parameter efficiency, and computational complexity. To ensure a fair comparison, we re-implemented all baseline methods under the same experimental conditions as TD3Net. 
Experiments marked with $\dagger$ differed only in the backend architecture, which adhered to the official code and hyperparameters released by the original authors. 
All models shared the same frontend architecture comprising 3D convolutions followed by ResNet-18.

Floating-point operations (FLOPs) are reported in gigaflops (GFLOPs), following the standard convention that each multiply-accumulate operation is counted as two floating-point operations.
FLOPs were calculated using an input sequence of shape (Batch=1, Channels=1, Frames=29, Height=88, Width=88) for the frontend, while the backend receives features of shape (Batch=1, Channels=512, Frames=29).
The frontend accounted for 16.9 GFLOPs---a relatively high value compared to the backend---primarily because it performs frame-wise processing after reshaping the input from $(B, C, T, H, W)$ to $(B{\times}T, C, H, W)$.

Although DC-TCN achieved higher accuracy (87.92\%) compared with BiGRU (85.35\%) and MS-TCN (86.90\%), it significantly increased backend complexity (40.55M parameters, 2.94 GFLOPs).
By contrast, TD3Net consistently achieved better accuracy with fewer parameters and lower FLOPs.
Specifically, TD3Net-Base attained 89.36\% accuracy using only 18.69M parameters and 1.56 GFLOPs, outperforming all baseline methods in both efficiency and effectiveness.
TD3Net-Best further improved performance, reaching the highest accuracy of 89.54\% with a moderate computational cost compared to prior methods (31.89M parameters, 1.92 GFLOPs).
These results indicate that TD3Net effectively mitigates the blind spots found in existing methods, resulting in robust temporal representations with reduced computational demands.

\section{Discussion}
\label{sec:discussion}

\begin{table}[t]
\renewcommand{\arraystretch}{1.1}
\caption{Evaluation results of the proposed method comprising different numbers of TD3 ($B$) and TD2 blocks ($N$) on the LRW test dataset. Reported parameters reflect the entire model, including both frontend and backend architectures.}
\label{table:hyperparams_result0}
\centering
\resizebox{\linewidth}{!}{%
\begin{tabular}{llll}
\toprule
\# TD3 blocks (B) & TD2 blocks (N) & \# Params (M) & Accuracy (\%) \\
\midrule
\midrule
2 & 16 & 31.00 & 88.92$\pm$0.07 \\
4 & 10 & 30.75 (Base) & 89.36$\pm$0.09 \\
6 & 8 & 30.72 & 89.06$\pm$0.05 \\
8 & 6 & 30.52 & 88.08$\pm$0.06 \\
\bottomrule
\end{tabular}
}
\end{table}

\begin{table}[t]
\renewcommand{\arraystretch}{1.1}
\caption{Evaluation results of the proposed method comprising different growth rates ($k$) and numbers of TD2 blocks ($N$) on the LRW test dataset.} 
\label{table:hyperparams_result}
\centering
\resizebox{\linewidth}{!}{%
\begin{tabular}{lllll}
\toprule
\multirow{2}{*}{\# TD2 blocks (N)} & \multirow{2}{*}{Growth rate ($k$)} & \multirow{2}{*}{\# Params (M)} & \multirow{2}{*}{Accuracy (\%)} \\
& & & & \\
\midrule
\midrule
\multirow{3}{*}{6} & 36 & 20.42 & 87.78$\pm0.09$ \\
 & 48 & 26.50 & 88.27$\pm0.13$ \\
 & 60 & 34.12 & 89.15$\pm0.11$
\\
\midrule
\multirow{3}{*}{8} & 36 &  25.16 &  88.40$\pm0.09$\\
 & 48 & 34.60 & 89.15$\pm0.13$\\
 & 60 & 46.47 & 89.11$\pm0.07$
\\
\midrule
\multirow{3}{*}{10} & 36 & 30.75 (Base) & 89.36$\pm0.09$ \\
 & 48 & 44.23 (Best)& 89.54$\pm0.08$\\ & 60 & 61.21 & 89.30$\pm0.11$
\\
\bottomrule
\end{tabular}
}
\end{table}

\begin{table}[th]
\caption{Evaluation results of the proposed method comprising different numbers of multi-dilated TC layers ($L$) in each TD2 block and two types of dilation factor patterns on the LRW test dataset. The linear pattern indicates that the sub-TC layer corresponding to the $i$\textsuperscript{th} skip connection in the multi-dilated TC layer has $d_{i}$ = 2i, and the exponential pattern has $d_{i}$ = $2^{i}$. Note that the notation is adopted from Eq. (\ref{eq:TD2_block}).}
\label{table:hyperparams_result2}
\centering
\resizebox{\linewidth}{!}{%
\begin{tabular}{lllll}
\toprule
\multirow{2}{*}{\# Layers (L)} & \multirow{2}{*}{Dilation factor pattern ($d_{i}$)} & \multirow{2}{*}{\# Params (M)} & \multirow{2}{*}{Accuracy (\%)} \\
& & & & \\
\midrule
\midrule
\multirow{2}{*}{4} & linear & \multirow{2}{*}{25.87} &  88.28$\pm0.14$\\
 & exponential &&  88.56$\pm0.09$\\

\midrule
\multirow{2}{*}{5} & linear &  \multirow{2}{*}{30.75 (Base)}&  89.03$\pm0.06$\\
 & exponential & & 89.36$\pm0.09$\\
\midrule

\multirow{2}{*}{6} & linear & \multirow{2}{*}{36.27} &  89.15$\pm0.13$\\
 & exponential & & 89.37$\pm0.09$\\

\bottomrule
\end{tabular}
}
\end{table}

\subsection{Impact of Hyperparameters}
\label{sec:determining hyperp}
In this section, we empirically identify the impact of different components of TD3Net and report the best settings for the LRW dataset. 
The proposed method includes four primary hyperparameters: the number of TD3 blocks (B), number of TD2 blocks (N), number of TC layers (L), and growth rate ($k$).
Increasing these hyperparameters significantly raises the complexity of the model.
Therefore, to ensure that the complexity level remains consistent with that of existing methods, appropriate values for these hyperparameters were determined. 
Moreover, given the wide range of potential hyperparameter combinations, conducting experiments with all options is computationally expensive. 
Thus, we employed a systematic approach, narrowing the options step-by-step, as indicated in Tables~\ref{table:hyperparams_result0}-\ref{table:hyperparams_result2}.
During this process, we empirically selected candidate values for hyperparameters that had not yet been assigned values.

First, we conducted an experiment to determine the number of TD3 blocks under similar complexity conditions. The best results were obtained with four TD3 blocks, as reported in Table~\ref{table:hyperparams_result0}. According to the results, we set $B$ to 4 in all subsequent experiments.
Then, we evaluated the effect of the number of TD2 blocks ($N$) in each TD3 block and the growth rate ($k$). 
In these experiments, the networks were trained by varying $N = $ \{6, 8, 10\} and $k = $ \{36, 48, 60\}, while maintaining the number of layers (L) in each TD2 block at 5. Other hyperparameters were set as described in Section~\ref{sec:arch_details}. The experimental results are presented in Table~\ref{table:hyperparams_result}. When comparing the accuracy of the models, it was found that increasing $N$ was more efficient than increasing $k$ for improving performance in terms of model size. Specifically, the results obtained using similar configurations \{$N =$ 8, $k =$ 48\} and \{$N =$ 10, $k =$ 36\} demonstrated the efficiency of $N$. In addition, the results of models with \{$N =$ 10, $k =$ 48\} and \{$N =$ 10, $k =$ 60\} revealed that over-parameterization degraded the quality of the model. Based on these findings, $N$ = 10 and $k$ = 36 were selected as the hyperparameters for subsequent experiments.
Subsequently, to explore the appropriate receptive field range of the network, we further evaluated the effects of different $L$ = \{4, 5, 6\} and two types of dilation factor patterns $d_{i}$, including linear and exponential patterns. 
Among these, the linear pattern indicated that the sub-TC layer corresponding to the $i$\textsuperscript{th} skip connection in the multi-dilated TC layer had $d_{i}$ = $2i$, while the exponential pattern had $d_{i}$ = $2^{i}$.
For example, networks with $L$ = 4 and $d_{i}$ = $2^{i}$ indicated that each TD2 block had four multi-dilated TC layers, each of which had dilation factors of \{\{1\}, \{1,2\}, \{1,2,4\}, and \{1,2,4,8\}\}, respectively. The results listed in Table~\ref{table:hyperparams_result2} show that the exponential pattern consistently outperformed the linear pattern for all values of $L$. However, when $L$ was increased to 6, in which case the maximum dilation of the linear and exponential patterns was 32 and 10, respectively, the performance gain was relatively lower compared with that when $L$ was increased to 5. This result implies that adding convolution with a dilation factor larger than the sequence length (LRW=29) does not enhance performance.

\subsection{Ablation Study}
\subsubsection{Limitations of TD2 Block in Isolation}
\label{sec:td2_limitation}
To independently evaluate the effectiveness of the TD2 block, we conducted a comparative experiment by replacing the backend architecture of TD3Net with two alternative configurations:
\begin{itemize}
  \item Dense-TCN: A baseline model composed of four dense blocks, each densely connected with dilated temporal convolution (TC) layers. The block depths are 6, 12, 24, and 16 TC layers, respectively. Within each block, the dilation factor for the $i$-th TC layer, where $i$ starts from 0, follows the increasing pattern $2 \times (i \bmod 8) + 1$. The dense blocks in this model correspond to graph (a) in Fig.~\ref{fig:TMDB}.
  \item TD2Net: A TD3Net variant composed of four TD2 blocks, where each block is densely connected with multi-dilated TC layers. The only difference from Dense-TCN is the use of multi-dilated TC layers, with each layer assigned a dilation factor following the increasing pattern $2 \times (i \bmod 8) + 1$. The TD2 blocks in this model correspond to graph (b) in Fig.~\ref{fig:TMDB}.
\end{itemize}

\begin{table}[t]
\centering
\footnotesize
\caption{Ablation results on the TD2 block structure evaluated on the LRW test dataset. The dilation factor of (A) is identical to the maximum dilation value of the corresponding layer from (B).}
\label{tab:ablation_td2_structure}
\resizebox{\linewidth}{!}{%
\begin{tabular}{lllll}
\toprule
 & Configuration & \# Params (M) & Accuracy (\%) \\
\midrule
(A) & Dense-TCN & 30.23 & 87.62$\pm$0.10 \\
(B) & TD2Net    & 30.36 & 87.18$\pm$0.09 \\
\bottomrule
\end{tabular}%
}
\end{table}

Table~\ref{tab:ablation_td2_structure} shows that, despite incorporating multi-dilation to capture richer temporal information, TD2Net performed worse than Dense-TCN. In TD2Net, each block applies distinct dilation factors to the skip-connected features, aiming to extract multi-scale temporal patterns in parallel.

However, the structural complexity introduced by aggregating multi-dilated features across most layers can destabilize gradient flow during early training, impairing the model’s ability to capture salient temporal cues. In practice, removing the dense connections between TD2 blocks in TD3Net-base resulted in unstable training and convergence failure. Training stability was restored only after restructuring the network into four TD2 blocks with progressively increasing depths, inspired by DenseNet-121 \cite{huang2017densely}. This configuration facilitates gradual feature expansion and encourages efficient feature reuse while maintaining manageable parameter growth.

By contrast, Dense-TCN, with standard dilated TC layers, maintains structural simplicity and more coherent gradient flow, naturally offering easier optimization. These results suggest that module-level (TD2 block) design alone is insufficient to fully realize the expressive capacity of the network. To efficiently exploit structurally complex and expressive components such as TD2 blocks, incorporating additional dense connections across blocks is crucial. Such higher-level connections promote stable temporal aggregation and robust feature reuse, essential for effective temporal representation learning.

\begin{figure*}[ht]
\centering
\centerline{\includegraphics[scale = 0.51]{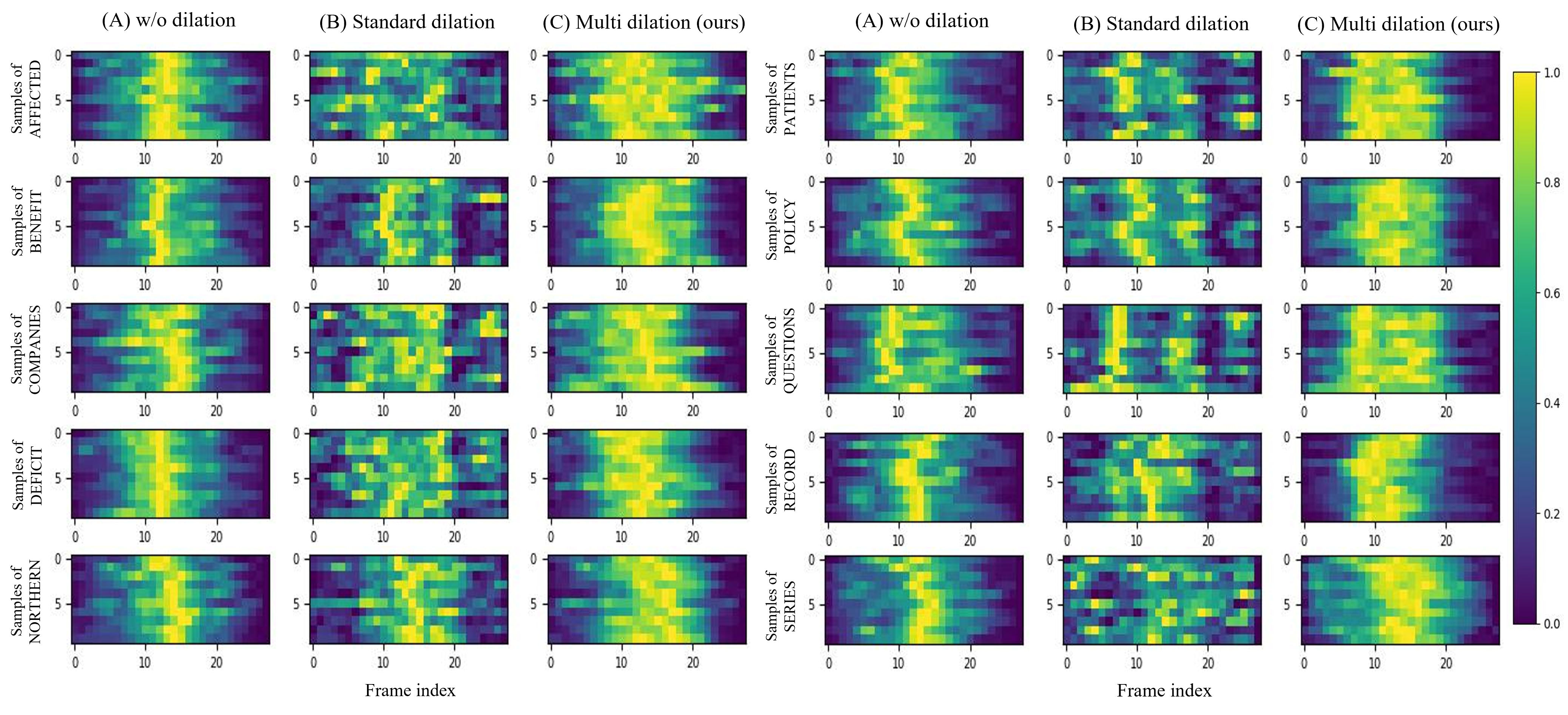}}
\caption{Activation map illustrating the magnitude ($l_2$ norm) of temporal features extracted using the backend model with different dilation methods. Our visualization was performed on ten randomly chosen samples from each of the ten randomly selected classes from the LRW test dataset.}
\label{fig:amap}
\end{figure*}

\subsubsection{Analysis of Multi-Dilation and TD2 Block Integration}
\label{sec:multi_dilation_effect}
This section compares three configurations to analyze how dilation strategies applied to skip connections influence temporal modeling, through both quantitative and qualitative evaluations. Specifically, we investigated the impact of blind spots that may arise from integrating dilation into skip connections and assessed whether the proposed multi-dilated TC layers can effectively mitigate these issues. To this end, the multi-dilated TC layers were replaced with non-dilated TC layers and with standard TC layers that use a single dilation factor, which may introduce blind spots. All configurations were designed with comparable parameter counts to ensure a fair comparison. Experimental results are summarized in Fig.\ref{fig:amap} and Table\ref{tab:ablation_TMDB}, with configurations (B) and (C) corresponding to graphs (a) and (b) in Fig.~\ref{fig:TMDB}, respectively.
Furthermore, motivated by the structural complexity and training instability of TD2 blocks discussed in Section~\ref{sec:td2_limitation}, we also investigated the necessity of dense connections between TD2 blocks. This analysis, informed by both previous and current experimental results, offers further insight into the nested dense connection design of TD3Net.

\paragraph{Qualitative Comparison} Fig.~\ref{fig:amap} visualizes the L2 norm for the final TD3 block output $\in \mathbb{R}^{C \times T}$ along the temporal axis, highlighting the time steps attended by the network when predicting the target utterance. The observed activation patterns in each configuration are summarized as follows:
\begin{itemize} 
\item (A) Non-dilated TC: Activations exhibit localized patterns confined to specific temporal regions, indicating limited temporal coverage.
\item (B) Standard TC: Compared to (A), activations extend over a wider temporal range but exhibit discontinuities. Therefore, using dilated TC layers in skip connections expands coverage, but identical dilation factors may introduce blind spots that compromise temporal coherence.
\item (C) Multi-dilated TC: Compared to (B), activations remain widely distributed across the temporal axis with improved continuity. Therefore, blind spots disrupt temporal consistency, which the multi-dilation strategy addresses by assigning different dilation factors to skip-connected features.
\end{itemize}

\begin{table}[t]
\caption{Ablation results on the impact of multi-dilation evaluated on the LRW test dataset. The dilation factor of (B) is identical to the maximum dilation value of the corresponding layer from (C).}
\label{tab:ablation_TMDB}
\centering
\resizebox{\linewidth}{!}{%
\small{
\begin{tabular}{lllll}
\toprule
 & Configuration & \# Params (M) & Accuracy (\%) \\
\midrule
(A) & w/o dilation & 30.75 & 87.83$\pm0.10$\\
(B) & Standard dilation & 30.75 & 88.62$\pm0.13$ \\
(C) & Multi dilation (ours) &  30.75 & 89.36$\pm0.09$  \\
\bottomrule
\end{tabular}
}}
\end{table}

\paragraph{Quantitative Comparison}
To better understand the effectiveness of the proposed multi-dilated TC layers, we analyzed how the observed activation patterns relate to the performance differences reported in Table~\ref{tab:ablation_TMDB}. 
Configuration (B) improved accuracy from 87.83\% in (A) to 88.62\%, underscoring the effectiveness of capturing broader temporal ranges. Configuration (C) achieved a further gain, increasing accuracy from 88.62\% to 89.36\% over (B), indicating the importance of maintaining continuity alongside broader temporal coverage. These results suggest that blind spots interfere with modeling temporal dependencies by disrupting continuity. The proposed multi-dilation strategy alleviates this issue, enabling more consistent integration of temporal representations and ultimately leading to significant improvements in lipreading performance.

Meanwhile, while configuration (B) is based on Dense-TCN’s dense block structure and configuration (C) on TD2Net's TD2 block, the current results differ from previous experiments where Dense-TCN outperformed TD2Net.
Configuration (C) outperformed (B), demonstrating that within a nested dense connection structure, multi-dilated layers are more effective than standard dilated layers. These findings highlight that effectively exploiting the expressive capacity of the TD2 block requires structural support via dense integration across blocks, thereby emphasizing the necessity of the TD3 block.

\subsubsection{Summary of Findings}
The experimental results confirm that multi-dilated layers within the TD2 block effectively mitigate blind spots induced by standard dilation, enabling more comprehensive temporal information capture. Despite the high representational capability of the TD2 block, it exhibits inherent limitations in terms of training stability and overall performance when used in isolation. These findings suggest that simply employing multi-dilated layers is not sufficient; careful architectural design---such as that of TD3Net---is essential to fully harness their potential. To address these limitations, TD3Net incorporates additional dense connections between TD2 blocks, enhancing temporal information integration. This nested integration approach substantially leverages the representational power of TD2 blocks and has been empirically shown to improve lipreading performance.

\subsection{TD3Net for Real-Time Lipreading and AVSR Systems}
Most existing lipreading studies have primarily focused on improving recognition accuracy, often overlooking the practical constraints associated with real-time deployment \cite{sheng2024deep}. For example, recent studies tend to adopt computationally intensive temporal architectures such as transformer, Conformer \cite{gulati2020conformer}, and Branchformer \cite{peng2022branchformer}, which prioritize performance over inference efficiency.
However, as lipreading systems advance toward real-world applications, the demand for real-time operation is expected to grow. In this regard, the proposed TD3Net is well-suited for such scenarios, as it is based on TCNs that offer several efficiency advantages, such as relatively low computational complexity and a causal design that is particularly beneficial for streaming or low-latency applications.

To illustrate the computational differences between temporal modeling architectures, consider a 1D convolution layer with a kernel size of $k = 3$, which has a computational cost of $3NC^2$, where $N$ is the sequence length and $C$ is the channel dimension. By contrast, a self-attention layer includes three linear projections to compute the query, key, and value matrices, each with a cost of $NC^2$, and an attention operation with a cost of $N^2C$. The total computational cost of the self-attention layer is therefore approximately $3NC^2 + N^2 C$. This gap becomes especially problematic in sentence-level lipreading, where the input sequence is typically much longer than in word-level tasks. As $N$ increases, the quadratic term $N^2C$ dominates, leading to substantial overhead in real-time or resource-constrained environments. In such cases, lightweight alternatives such as TCN-based models offer a more practical and scalable solution for temporal modeling.

Furthermore, the use of standard self-attention in real-time lipreading scenarios presents an additional challenge related to causality. Most self-attention mechanisms are noncausal by default, meaning they attend to both past and future frames. While this is beneficial for offline recognition, it is incompatible with real-time inference, where future frames are unavailable. To enable streaming applications, causal (masked) self-attention must be employed. However, if the model is trained with full self-attention but deployed with causal masking at inference time, a significant training–inference mismatch occurs. This mismatch can lead to noticeable performance degradation, as the model learns to rely on future context during training that is no longer accessible at test time. By contrast, TD3Net is designed to support causal processing during both training and inference, enabling consistent behavior without training-inference mismatch. This property is essential for stable and reliable real-time lipreading.

This robustness in real-time settings makes TD3Net a promising candidate not only as a standalone temporal model but also as a complementary module in broader lipreading and audiovisual speech recognition (AVSR) systems. As mentioned above, recent lipreading and AVSR frameworks commonly employ Conformer and Branchformer architectures to jointly model global and local temporal dependencies through a combination of self-attention and convolution. However, their reliance on standard convolution blocks with fixed dilation restricts the diversity of temporal patterns that can be captured. Replacing these blocks with TD3Net’s multi-dilated convolution blocks could enhance temporal modeling by enabling the extraction of more diverse and flexible patterns across varying time scales, potentially improving performance in both visual-only and multimodal lipreading systems.

\section{Conclusion}
\label{sec:conclusion}
In this study, we proposed TD3Net to maintain a wide and dense receptive field without blind spots for word-level lipreading, which is crucial for recognizing individual words and serves as the basis for recognizing longer sequences. TD3Net, which combines nested dense skip connections and multi-dilated TC layers, can learn multi-temporal representations in almost all layers without losing information about the continuity of lip movements induced by blind spots. The experimental results on two public benchmark datasets revealed that the proposed approach achieves performance comparable to existing state-of-the-art methods. In particular, experiments conducted under tightly controlled conditions demonstrated that the proposed method yielded superior performance compared with other backend architectures while maintaining lower computational cost and fewer parameters. Moreover, we visually observed that standard dilation in dense skip connections created blind spots that disrupt temporal continuity, while our multi-dilation strategy preserved it. This capability is achieved in TD3Net through the dense integration of TD2 blocks, which stabilizes training and enhances temporal modeling, ultimately leading to significant improvements in lipreading performance. Although the proposed method demonstrated excellent lipreading performance, this study focused solely on recognizing individual words; therefore, further research is needed to develop a method that can accurately recognize speech at the sentence level in lipreading.
\bibliographystyle{elsarticle-num-names}
\bibliography{paper}

\appendix

\section{Model Implementation} \label{sec:appendixa}
Algorithm \ref{algo:a1} presents the PyTorch-style pseudocode for TD3Net and provides a simple and comprehensive illustration of the algorithm's implementation. The pseudocode includes abstractions and simplifications designed to enhance clarity. The algorithm is a simplified version of an actual implementation.

\definecolor{commentcolor}{RGB}{64,140,128}   
\newcommand{\PyCode}[1]{\bfseries\footnotesize\ttfamily\textcolor{black}{#1}} 
\newcommand{\PyComment}[1]{\bfseries\footnotesize\ttfamily\textcolor{commentcolor}{\# #1}}  

\begin{algorithm*}[ht]
\caption{PyTorch-style pseudocode for TD3Net} \label{algo:a1}
\SetAlgoLined
    \PyComment{input\_tensor: Input tensor with shape (bs: Batch size, ch: The number of channels, T: Sequence length, H: Height, W: Width)} \\
    \PyComment{output\_tensor: Output tensor with shape (bs: Batch size, ch: The number of classes)}\\
    \PyCode{} \\
    
    \PyComment{Hyperparameters}\\
    \PyComment{B: The number of TD3 blocks}\\
    \PyComment{N: The number of TD2 blocks}\\
    \PyComment{L: The number of multi-dilated TC layers in a TD2 block}\\
    \PyComment{k: The number of channels generated by each layer of the multi-dilated TC layer}\\
    \PyComment{c: Compression ratio that reduces the output channels of each TD3 block}\\
    \PyComment{t: Transition ratio that reduces the output channels of each TD2 block}\\
    \PyComment{bc: The number of channels produced by the first layer of the TD2 block}\\
    \PyCode{} \\

    \PyComment{Note} \\
    \PyComment{To ensure code readability, all function definitions are designed to take both tensors and hyperparameters as input parameters.} \\ 
    \PyComment{Conv1D(out\_c)(x): Forward pass of a layer to compute an output with channel size out\_c for tensor x.} \\
    \PyComment{rearrange(x): A reshape operation to transform the shape of tensor x.} \\
    \PyComment{f(): A two-dimensional convolutional network with spatial average pooling (e.g., ResNet-18).} \\
    \PyCode{} \\
    
    \PyCode{def TD2block(x, L, k, t, bc):} \\
    \Indp
        \PyComment{Define a function Multi-dilated TC layer (i.e., Eq. (\ref{eq:TD2_block}))} \\
        \PyCode{def MDlayer(x, L, k, bc):}\\
        \Indp
            \PyCode{md\_out = Conv1D(out\_c=k, dilation=1)(x[:, :bc])} \PyComment{Apply initial Conv1D} \\ 
            \PyCode{for i in range(L):} \\
            \Indp
                \PyCode{md\_out += Conv1D(out\_c=k, dilation=2**(i+1))(x[:, bc+(i*k):bc+(i+1)*k])} \\
            \Indm
            \PyCode{return md\_out} \\
        \Indm
        \PyCode{}\\
        \PyComment{Apply a bottleneck (bn) layer comprising a Conv1D with a kernel size of 1} \\
        \PyCode{x = bnLayer(out\_c=bc)(x)} \\
        \PyCode{for i in range(L):}\\
        \Indp
            \PyCode{x = Concat([x, MDlayer(x, i, k, bc)], dim=1)} \PyComment{Concatenation in the channel dimension} \\
        \Indm
        \PyCode{td2\_out = bnLayer(out\_c=L*k*t)(x)} \\
        \PyCode{} \\
        \PyCode{return td2\_out} \\
    \Indm
    \PyCode{}\\
    \PyCode{def TD3block(x, N, L, k, c, t, bc):} \\
    \Indp   
        \PyComment{Apply TD2blocks N times} \\
        \PyCode{for \_ in range(N)} \\ 
        \Indp   
            \PyCode{x = Concat([x, TD2block(x, L, k, t, bc)], dim=1)}\\
        \Indm 
        \PyCode{td3\_out = bnLayer(out\_c=x.size(dim=1)//c})(x)  \\ 
        \PyCode{} \\
        \PyCode{return td3\_out} \\
    \Indm 
    \PyCode{}\\

    \PyCode{def TD3Net(input\_tensor, B, N, L, k, t, bc):} \\
    \Indp   
        \PyCode{bs, ch, T, H, W = input\_tensor.size()} \\
        \PyCode{}\\
        \PyComment{Apply a frontend model} \\
        \PyCode{x = Conv3D(out\_c=64)(input\_tensor)} \\
        \PyCode{x = rearrange(x, 'bs 64 T H W -> (bs T) 64 H W')} \\
        \PyCode{x = ResNet(out\_c=512)(x)} \PyComment{ResNet with spatial pooling layer}\\ 
        \PyCode{x = rearrange(x, '(bs T) 512-> bs 512 T')} \\
        \PyCode{}\\
        \PyComment{Apply a backend model} \\
        \PyCode{for i in range(B)} \\ 
        \Indp   
            \PyCode{c = 1 if i == (B - 1) else 0.5} \\
            \PyCode{x = TD3block(x,N,L,k,c,t,bc})(x)\\
        \Indm 
        \PyCode{}\\
        \PyComment{Apply a Classification layer} \\
        \PyCode{x = torch.mean(x, dim=-1)} \\
        \PyCode{output\_tensor = Classifier()(x)} \\
        \PyCode{} \\
        \PyCode{return output\_tensor} \\
    \Indm 
    \PyCode{}\\    
    
\end{algorithm*}

\end{document}